\documentclass[a4paper,twoside]{article}

\usepackage{epsfig}
\usepackage{subcaption}
\usepackage{calc}
\usepackage{amssymb}
\usepackage{amstext}
\usepackage{amsmath}
\usepackage{amsthm}
\usepackage{multicol}
\usepackage{pslatex}
\usepackage{apalike}
\usepackage{hyperref} 
\DeclareMathOperator*{\argmax}{arg\,max}

\usepackage{SCITEPRESS}     

\begin{document}

\title{Quantifying Multimodality in World Models}

\author{\authorname{Andreas Sedlmeier, Michael Kölle, Robert Müller, Leo Baudrexel and Claudia Linnhoff-Popien}
\affiliation{LMU Munich, Munich, Germany}
\email{andreas.sedlmeier@ifi.lmu.de}
}


\abstract{
Model-based Deep Reinforcement Learning (RL) assumes the availability of a model of an environment's underlying transition dynamics.
This model can be used to predict future effects of an agent's possible actions.
When no such model is available, it is possible to learn an approximation of the real environment, e.g. by using generative neural networks, sometimes also called World Models.
As most real-world environments are stochastic in nature and the transition dynamics are oftentimes multimodal, it is important to use a modelling technique that is able to reflect this multimodal uncertainty.
In order to safely deploy such learning systems in the real world, especially in an industrial context, it is paramount to consider these uncertainties.
In this work, we analyze existing and propose new metrics for the detection and quantification of multimodal uncertainty in RL based World Models.
The correct modelling \& detection of uncertain future states lays the foundation for handling critical situations in a safe way, which is a prerequisite for deploying RL systems in real-world settings.
}

\onecolumn \maketitle \normalsize \setcounter{footnote}{0} \vfill

\section{\uppercase{Introduction}}
\label{sec:introduction}
While model-free reinforcement learning (RL) has produced a continuous stream of impressive results over the last years \cite{mnih2015human} \cite{vinyals2019grandmaster},
interest in model-based reinforcement learning has only recently experienced a resurgence \cite{alphazero} \cite{schrittwieser2020mastering}.
Although at first, approaches like World Models \cite{ha18} might seem more complex, they also promise to tackle some important aspects like sample-efficiency, that still hinder wide deployment of RL in the real world.
Besides the potential benefits, it is still necessary to consider non-functional aspects when developing model-based RL systems.
Most important might be guaranteeing reliable behaviour in uncertain conditions.
Especially considering industrial systems, such uncertainty could lead to potential safety risks when wrong predictions of the learning system lead to the execution of harmful actions.
Consequenly, considering and in the case of model-based RL, correct modelling of the present uncertainty is of utmost importance in order to build reliable systems.

Existing work in this area has mostly focused on differentiating between aleatoric and epistemic uncertainty, and the question of which kind is relevant to certain tasks \cite{kendall2017uncertainties} \cite{osbandPrior}.
The work at hand focuses on the uncertainty's aspect of multimodality.
Considering the goal of learning a model for model-based RL, this kind of uncertainty arises whenever the distribution of the stochastic transition dynamics of the underlying Markov decision process (MDP) is multimodal.
On one hand, using a modelling technique which is not able to reflect this multimodality would lead to a suboptimal model.
On the other hand, assuming a modelling technique which is able to correctly reflect this multimodality, being able to detect
the presence of states with multimodal transition dynamics would be of great value.
If one is able to detect these uncertain dynamics, it becomes possible to guarantee robustness, for example by switching to a safe policy or handing control over to a human supervisor \cite{amodei2016concrete}.
In this work, we analyze existing and propose new metrics for the detection and quantification of multimodal uncertainty in World Model architectures.
We begin, by introducing basic preliminaries in the next section, followed by related-work in section 3. 
Section 4 explains the basic concepts underlying the quantification of multimodality and introduces existing and new metrics which will be used for evaluation.
We explain our experimental setup in section 5, followed by evaluation results in section 6 and a conclusion in section 7.

\section{\uppercase{Preliminaries}}
In this section we discuss the necessary preliminaries our work builds upon. We start with a brief introduction to reinforcement learning (RL) and World Models~\cite{ha18}, a model-based RL architecture that our work makes heavy use of. Next, we introduce Mixture Density Networks and review uncertainty and multimodality.

\subsection{Model-Based RL \& World Models}
In RL, an agent interacts with its environment to maximize reward. The environment is formally specified in terms of a Markov decision process (MDP).
An MDP is a tuple $(S, A, P, R, \gamma)$ where $S$ is the set of states and $A$ the set of actions Let $s_t \in S$ and $a_t \in A$ be the state and action at timestep $t$, then $P(s_{t+1} \vert s_t, a_t)$ denotes the transition function, i.e., the dynamics of the environment, $R: S \times A \rightarrow \mathbb{R}$ the reward function and $\gamma \in (0,1)$ is the discount factor. The goal is then to find a policy $\pi^*: S \rightarrow A$ which maximizes the following objective:
$$
\pi^* = \argmax_{\pi} \, \mathbb{E}_{\pi} \bigg[ \sum_{t=0}^{\infty} \gamma^k \,R(a_{t}, s_{t}) \bigg]
$$
$\gamma$ is needed to make the  infinite sum converge and by further decreasing $\gamma$ one favors short time reward.
Model-free and model-based RL can be differentiated by the question of whether the agent has access to or learns a model of $P$ and $R$.
In the case of model-free RL, the agent can only interact directly with the environment via policy $\pi$ and receive $s_t$ and $r_t = R(a_t, s_t)$.
In model-based RL by contrast, the agent can plan by using the model to query possible future consequences of it's actions.
In both cases, interacting with the environment produces trajectories of the form $(s_t, a_t, r_t, s_{t+1}, a_{t+1}, r_{t+1} \dots)$.

World Models are a special case of model-based RL introcued in~\cite{ha18}.
In this architecture, no pre-supplied model is available and instead, the agent aims to learn a compressed spatial and temporal representation of the environment. Results show that a model learned this way can lead to improved sample efficiency in optimizing the RL policy.
Architectural details of the World Model architecture will be introduced in more depth in section \ref{subsubsec:algorithms}.

\subsection{Mixture Density Networks}
The idea of Mixture Density Networks (MDNs) was introduced in \cite{Bishop94}.
Summarized succinctly, the goal of their development was being able to solve a supervised learning task which has a non-Gaussian distribution.
Such cases often arise with so called inverse problems, where the distribution is multimodal.
Bishop presents the MDN as a flexible mixture model framework which can model arbitrary conditional densities.
In the case of using gaussian components, the conditional distribution $p(y|x)$ is calculated as:
\begin{equation}
    p(y|x) = \sum_{k=1}^{K} \pi_k(x) \mathcal{N}(y|\mu_k(x), \sigma^2_k(x))
\end{equation}
with $k$ being the amount of mixture components, $\pi_k(x)$ the mixture coefficients and $\mathcal{N}$ normal distributions with means $\mu_k(x)$ and variances $\sigma^2_k(x).$ 
With the exception of $k$ which has to be defined as a hyperparameter, these parameters of the mixture model can then be learned using any kind of neural network.

\subsection{Uncertainty and Multimodality}
In the field of machine learning, it is common to differentiate between two types of uncertainty. First, aleatoric uncertainty, which describes uncertainty inherent in the data, for example, due to measurement inaccuracy. The other is epistemic uncertainty, which can be described as uncertainty regarding the parameters or structure of the model. This kind of uncertainty can be reduced by more data, whereas aleatoric uncertainty is irreducible.\\
As a special kind of aleatoric uncertainty, multimodality refers to the shape of the aleatoric uncertainty's distribution.
If the distribution has a single mode, it is called unimodal, if it has at least two modes, it is called multimodal.
In the field of RL and MDPs, such multimodal uncertainty can, among others, be present if the state transition function $P(s_{t+1} \vert s_t, a_t)$ is stochastic in nature.

\section{\uppercase{Related Work}}

\subsection{Bump Hunting}
In the research field of statistics, the search for multimodality is sometimes called Bump Hunting.
Here, methods for low-dimensional data exist that try to detect the presence of multiple local maxima.
The Patient Rule Induction Method (PRIM) \cite{Friedman99} is a method frequently used in this field.
Given a dataset, PRIM reduces the search range iteratively until a subrange with comparatively high values is found.
If data is two- or multi-dimensional, PRIM may not be able to distinguish two different modes from each other \cite{Polonik10}.
As our work aims to detect multimodality in high dimensional, possibly image based data, these classical methods are not applicable.

\subsection{Uncertainty-based OOD Detection}
A slightly different line of work is concerned with detecting untrained, out-of-distribution (OOD) situations in RL.
That research is related to the work at hand, as the goal is to achieve this by developing multiple methods for quantifying an RL agent's uncertainty.
PEOC \cite{peoc20} for example uses the policy entropy of an RL agent trained using policy-gradient methods, to detect increased epistemic uncertainty in untrained situations.
UBOOD \cite{ubood20} by contrast is applicable to value-based RL settings and is based on the reducibility of an agent's epistemic uncertainty in it's Q-Value function.
Although the methods differentiate between aleatoric and epistemic uncertainty to detect OOD situations, multimodality is not a focus.

\section{\uppercase{Quantifying Multimodality}}
\label{sec:quantify_multimod_mdn}
As explained in the introduction, being able to detect multimodality is an important first step towards building reliable \& safe learning systems.
Consequently, in this section, we focus on the question of how to quantify multimodality.
We begin in \autoref{subsec:multimod_mdn} by analyzing the case of a simple 1-dimensional regression case, modelled using an MDN.
Section \ref{subsec:multimod_worldmodels} then follows up by introducing the more complex case of 
detecting multimodal state-transitions in a high-dimensional World Model setting.

\subsection{Multimodality in Mixture Density Networks}
\label{subsec:multimod_mdn}
\begin{figure}[htb]
    \centering
    \begin{subfigure}{.48\columnwidth}
        \includegraphics[width=\textwidth]{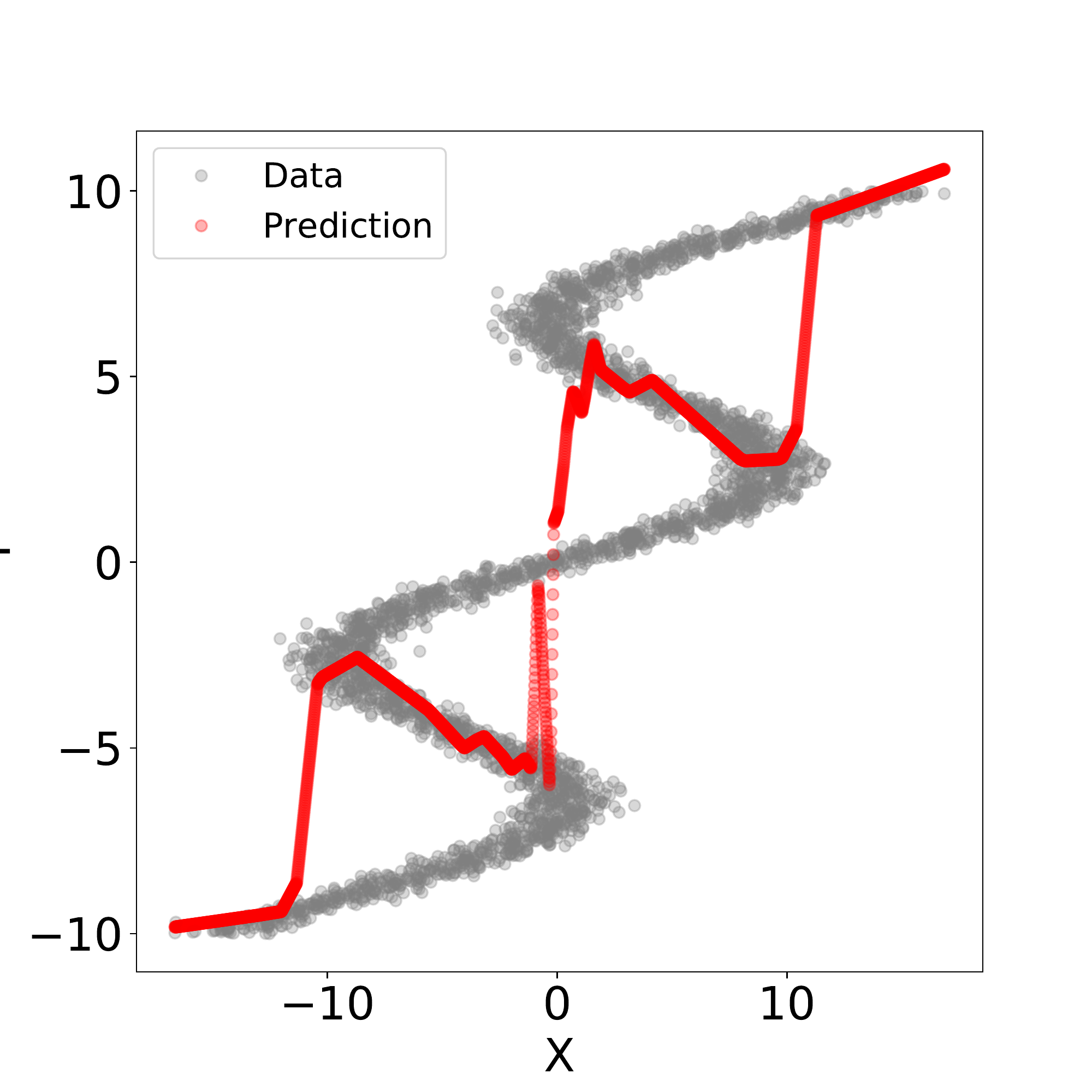}
        \caption{Deep NN - MSE}
        \label{fig:inv_sinus_pred_mlp}
    \end{subfigure}
    \begin{subfigure}{.48\columnwidth}
        \includegraphics[width=\textwidth]{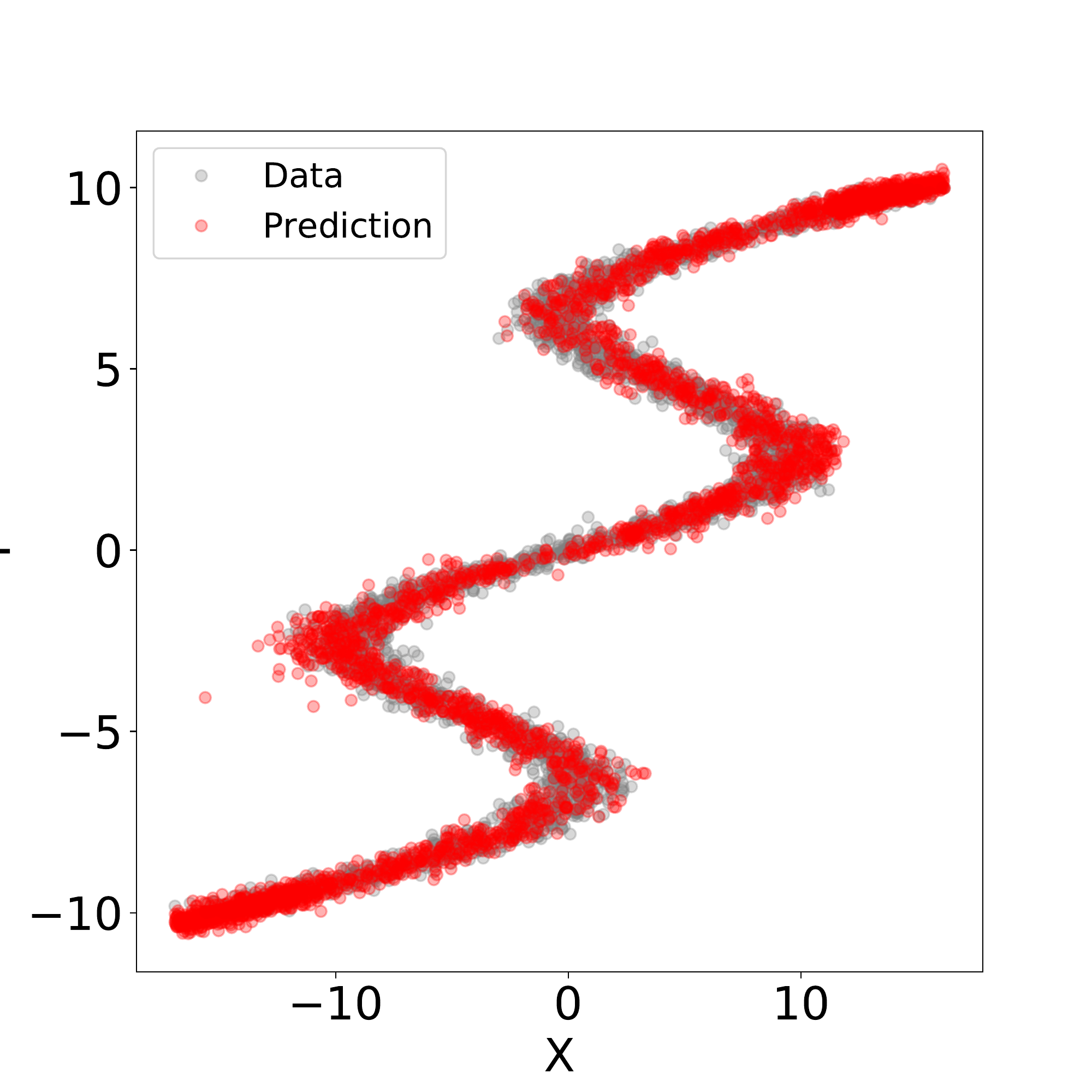}
        \caption{MDN}
        \label{fig:inv_sinus_pred_mdn}
    \end{subfigure}
    \caption{Predictions of (a) a simple deep neural network using MSE as the loss function, and (b) a MDN with $k=5$ components, trained on the inverse sine wave dataset.}
    \label{fig:inv_sinus_pred}
\end{figure}

We explain the following ideas using a simple synthetic dataset.
It is inspired by \cite{Bishop94} and is generated by inverting a noisy sine wave \footnote{Data is generated using: $f(x) = x + 7 * \sin{(0,7 * x)}$ and then adding noise from a standard normal distribution. The inverse problem is obtained by exchanging the roles of $x$ and $y$.}.
From the perspective of the work at hand, it is interesting as it contains both areas of unimodality as well as multimodality.
\autoref{fig:inv_sinus_pred} shows the failure of a simple deep neural network using MSE as the loss function, to correctly model the function $f(x) = y$. This is due to the fact, that the network tries to reduce the mean-squared error while only being able to make point predictions.
The predictions generated by a $5$ component MDN (red dots in \autoref{fig:inv_sinus_pred_mdn})
by contrast correctly approximate the target function.

\begin{figure}[htb]
    \centering
    \begin{subfigure}{.48\columnwidth}
        \includegraphics[width=\textwidth]{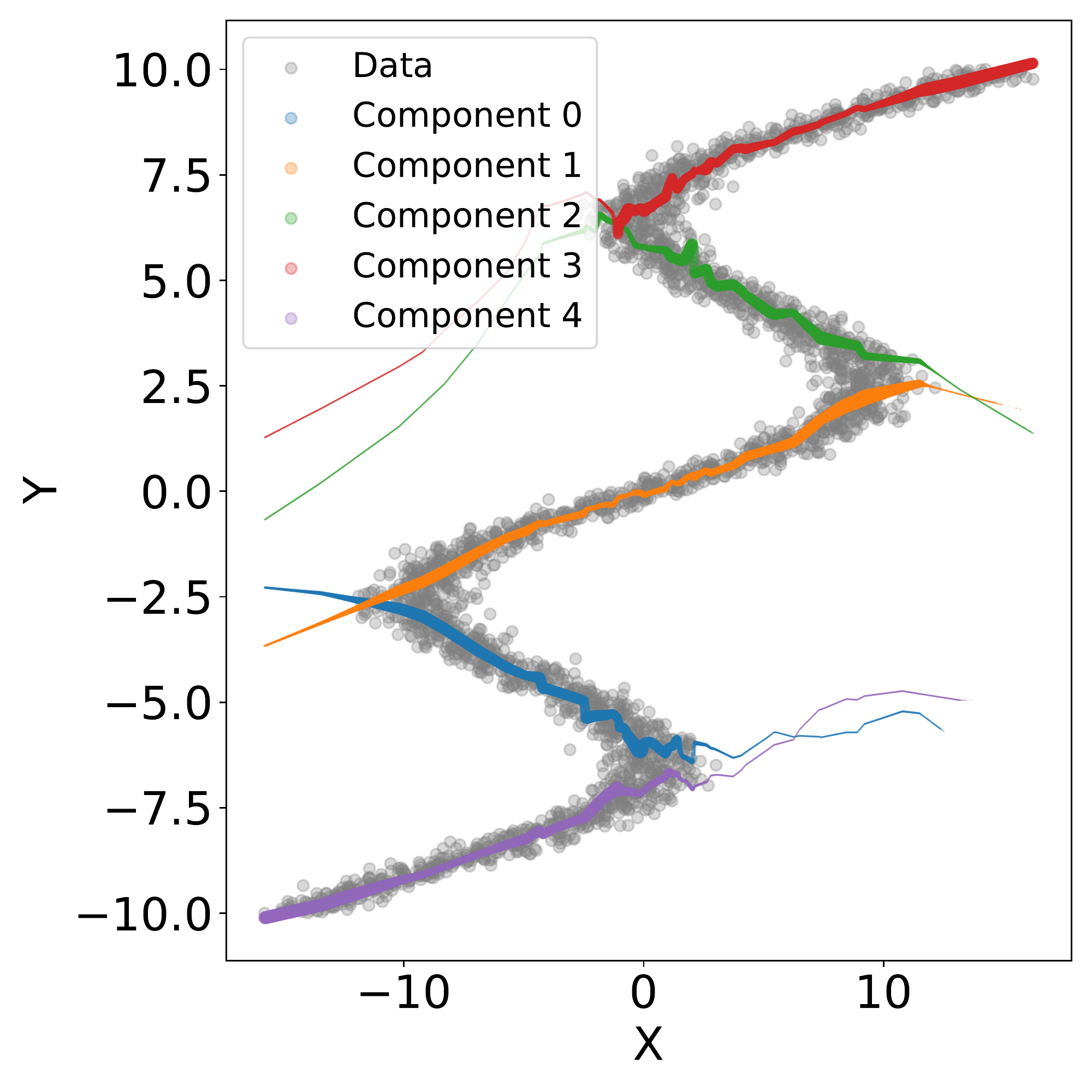}
        \caption{Mean prediction}
    \end{subfigure}
    \begin{subfigure}{.48\columnwidth}
        \includegraphics[width=\textwidth]{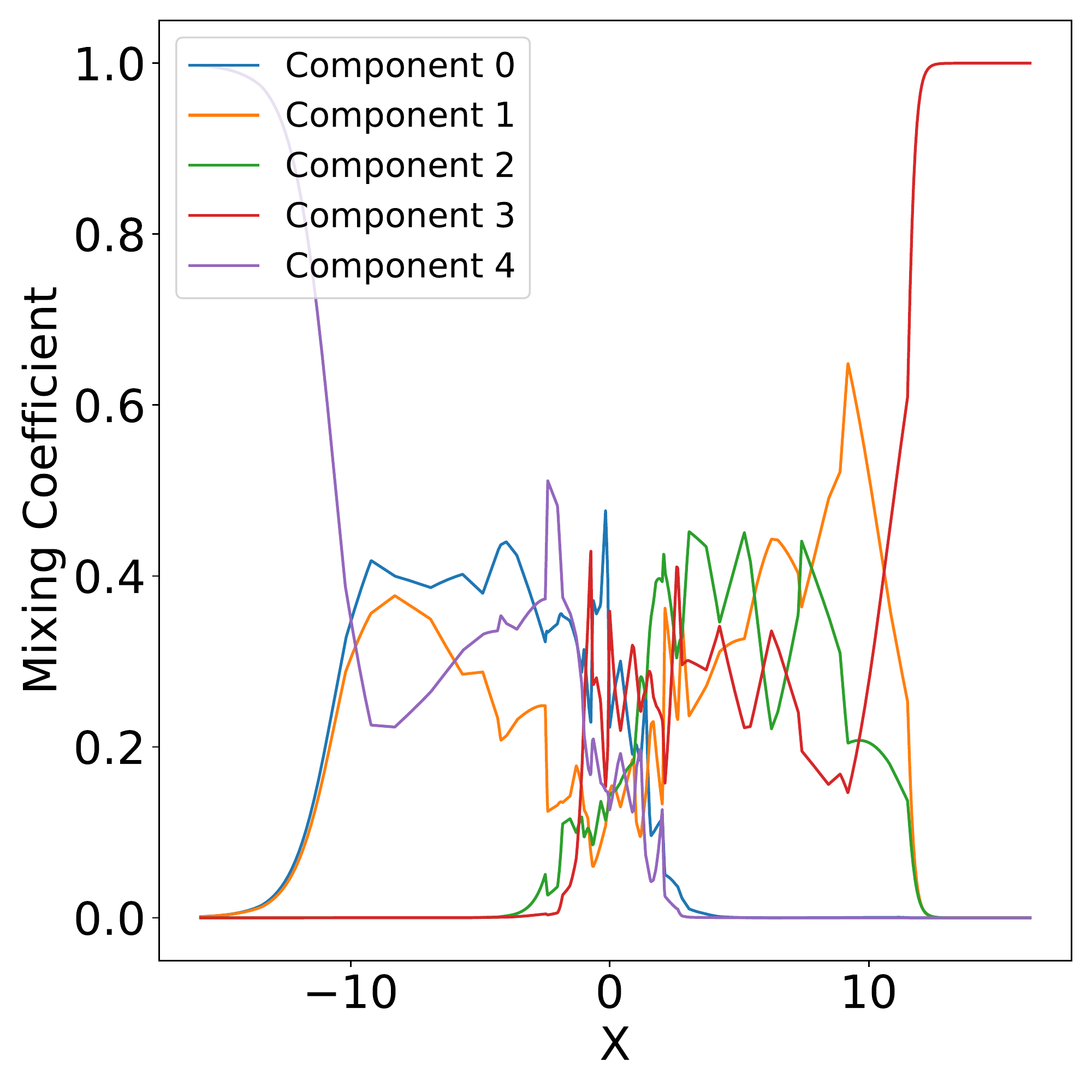}
        \caption{Mixing coefficients}
        \label{fig:inv_sinus_weights_mixing}
    \end{subfigure}
    \caption{MDN with $k=5$ components, fitted on the inverse sine wave dataset. Line width in (a) is calculated by multiplying the component's mixing coefficient $\pi$ with it's standard deviation.}
    \label{fig:inv_sinus_weights}
\end{figure}

When looking at the activations of the mixing coefficients of the trained MDN when predicting on this dataset, it possible to analyse how this is achieved.
Each color in \autoref{fig:inv_sinus_weights} 
represents a single component of the mixture model. Line width is calculated by multiplying the component's mixing coefficient $\pi$ with it's standard deviation.
Visualized this way, it becomes apparent that different components focus on different areas of the dataset.

Regarding the quantification of multimodality, \autoref{fig:inv_sinus_weights_mixing}
gives a first hint. Here, the mixing coefficients $\pi(x)$ of all components are visualized.
It becomes apparent, that in areas of increasing unimodality with $-10 > x > 10 $, a single component dominates. In areas of multimodality, multiple components contribute.

\subsection{Quantifying Multimodality in World Models}
\label{subsec:multimod_worldmodels}
While a visual analysis of the multimodality was possible in the simple 1D regression case presented in the previous section,
such an approach is no longer possible when using high dimensional input data. World models, as one of the most prominent representatives of model based RL, are most often based on input data in the form of high dimensional images. In the work of \cite{ha18} for example, input tensors have a size of $64x64x3 = 12288$ dimensions. Using any kind of classical multimodality detection technique from the related statistics literature directly on the input data is impossible in these dimensionalities.
Further more, the focus of our work is more complex than simply analyzing state inputs.
Instead, we focus primarily on quantifying multimodality in order to detect multimodal state-transitions.
Consequently, any multimodality quantification and detection which aims to do this, needs to be applied later on, in the so called Memory Model of the World Model pipeline. Here, future states are predicted and multimodal transitions can possibly be detected.
The exact process and integration point of the developed multimodality metrics will be described in section \ref{subsubsec:eval_pipeline}.

\subsection{Multimodality Metrics}
\label{subsec:multimod_metrics}
This section presents and discusses the different multimodality metrics that will be evaluated:
Two existing ones, SEMD and JSD, as well as two newly developed ones, we call MCE and WAKLD.

\subsubsection*{Mixing coefficient entropy}
A first, simple approach we propose is to compute the Shannon Entropy $H(X) = - \sum^n_{i=1}p(x_i)\log p(x_i)$ of an MDNs mixture coefficients for multimodality quantification. We call the resulting metric Mixing Coefficient Entropy (MCE). It is constructed as follows:

We interpret the $k$ mixing coefficients of a MDN as the categories of a multinomial distribution of size $k$, where each mixing coefficient's activation corresponds to a category's probability:  $\pi_i(x) = p_i$.

This is valid, as the mixing coefficients of a MDN must satisfy the constraint

\begin{equation*}
    \sum\limits^k_{i=1}\pi_i(x) = 1, \mkern42mu 0 \leq \pi_i(x) \leq 1
\end{equation*}
according to definition \cite{Bishop94}.
It is then possible to compute the entropy of this distribution.
In order to be able to compare entropy values of MDNs with different amount $k$ of components, it is helpful to normalize this value:
\begin{equation*}
    MCE(p_1,p_2,\dots,p_k) = \frac{H(\pi)}{H_{max}} = - \sum^k_{i=1} \frac{\pi_i(x)\log \pi_i(x)}{\log k},
\end{equation*}
with $p_i = \pi_i(x)$. This restricts the possible values to the range $[0,1]$.

It is important to note possible failure cases when using the entropy of a MDN's mixing coefficients as a multimodality metric.
Consider the extreme case of a $2$-component MDN, where both components model the same distribution, e.g. when using Gaussian components, they have the same $\mu$ and $\sigma$.
This would self-evidently produce a unimodal predictive distribution, even when both component's mixing coefficients are $>0$.
The computed MCE in this case would also be $>0$ and wrongly signal multimodality.

\subsubsection*{Weighted Average Kullback-Leibler Divergence (WAKLD)}
In order to overcome this limitation, we designed a new metric with the explicit goal of incorporating the mixture components' individual distributions.
We call this metric \textit{Weighted Average Kullback-Leibler Divergence} (WAKLD).
It is constructed by calculating for every component, the weighted Kullback-Leibler Divergence (KLD) to every other component. By summing these up and weighing by the initial component, a score for the complete mixture distribution is calculated.

\begin{equation*}
    WAKLD(p_1,p_2,\dots,p_k) = \sum\limits^k_{i=1}\left(\pi_i \sum\limits^k_{j=1}\pi_j D_{KL}\left(p_i||p_j\right)\right)
\end{equation*}

Strongly multimodal MDNs produce large WAKLD values, while unimodal MDNs possess a small WAKLD value.
Expressed informally, the intuition behind the "double-weighing" is that e.g., a divergence to a component with low mixing-coefficient ($\pi_j$) should have a low impact on the metric, as well as all divergences of a component with low mixing-coefficient ($\pi_i$) to any other component, irrespective of their mixing-coefficient.

\subsubsection*{Self Earth Mover’s distance (SEMD)}
The authors of \cite{osama19} propose a metric for quantifying multimodality in MDNs based on the Earth-Mover's-Distance (EMD).
EMD is also known as the Wasserstein metric and informally describes the amount of work needed to transform one distribution into another.
The authors apply this to MDNs with arbitrary amount of components, by computing the EMDs between the component with the largest mixing coefficient (primary mode) and all other components.
This in effect computes the EMD to convert the multimodal mixture into a unimodal distribution defined by the primary mode.
A large SEMD value consequently indicates strong multimodality, while small SEMD indicates unimodality.

\subsubsection*{Jensen-Shannon Divergence (JSD)}
The Jensen-Shannon Divergence (JSD) is another information-theoretic divergence measure, based on the Shannon Entropy.
First introduced by \cite{Lin91}, it constitutes a symmetrization of the Kullback-Leibler Divergence and can be understood as the total KL divergence to the average distribution $\frac{p+q}{2}$.
It can be extended to more than two, individually weighted distributions:

\begin{equation*}
    D_{JS}(p_1, p_2, \dots, p_n) = H \left( \sum_{i=1}^{n} \pi_i p_i \right) - \sum_{i=1}^{n} \pi_i H(p_i),
\end{equation*}

\noindent with $p_i$ being a distribution and $\pi_i$ the respective weight.

\section{\uppercase{Experimental Setup}}
\label{sec:exp_setup}

We now explain the experimental setup used to evaluate the toy example based on the inverse sine wave function,
as well as the environment, network-architecture and evaluation pipeline used for the world model experiment.

\subsection{Inverse Sine Wave}

Section \ref{sec:quantify_multimod_mdn} introduced a simple toy problem, for analyzing MDN behaviour: The inverse sine wave.
We generate a dataset containing $3000$ linearly spaced points in the interval $[-10, 10]$ using the function $f(x) = (x + 7 * \sin{(0,7 * x)})$ and then adding noise from a standard normal distribution.
By exchanging the roles of $x$ and $y$, the inverse problem is obtained.

We fit this dataset using a simple fully-connected, 3-Layer MDN with $k=5$, i.e. $5$ mixture components.
All neurons use ReLU as the activation function, with the exception of the output neurons. Here, no activation function is used on the $\mu$ neurons. To enforce positivity of the neurons outputting a component's variance,
We follow the work of \cite{Brando17} and compute the output as $\sigma(x) = \text{ELU}(1,x) + 1 + 1e^{-7}$.
This way, increased numerical stability is achieved, compared to using the simple exponential function, as suggested by \cite{Bishop94}.
Training is performed over $1000$ episodes by minimizing the negative log-likelihood.
As fitting an MDN is inherently stochastic in nature, due to e.g. the random initialization of neural network weights and random data batching, we repeat this process for $50$ separate runs.
Based on these fitted networks, we then evaluate the four multimodality metrics introduced in \autoref{subsec:multimod_metrics}.

\subsection{World Model}

The following section presents the experimental setup used to evaluate the multimodality metrics presented in \autoref{subsec:multimod_metrics} in a high-dimensional world model setting.
The goal here is to differentiate multimodal state-transitions from unimodal ones.

\subsubsection{Environment}
Currently, there are no existing environments for benchmarking multimodality in deep RL.
Therefore, we chose to use an established RL benchmarking environment with no inherent stochasticity, i.e. only unimodal state-transitions: Coinrun \cite{procgen19}, a simple 2D platformer.
Multimodality can then be artificially introduced via action masking, as will be explained in more depth in \autoref{subsubsec:eval_pipeline}.
It is important to note here that the environment's state and transition dynamics only represent the foundation, based on which a generative model is learned, according to the world model architecture.
This basic setup allows us to generate a benchmark data-set with known ground-truth (multimodal/unimodal) state-transitions.
The exact pipeline that realizes this will be explained in section \ref{subsubsec:eval_pipeline}.

\subsubsection{Algorithms and Network Architectures}
\label{subsubsec:algorithms}
Our network architecture consists of the first two parts that make up the World Model as described in \cite{ha18}.
The Vision Model, which reduces a high-dimensional observation to a low-dimensional latent vector and the Memory Model, which makes predictions about future encodings based on past information.
We omit the third part, the controller, as optimizing a RL policy is not the focus of our work.
We use a convolutional VAE for the Vision Model that compresses each 2D-frame from the game to a smaller latent representation $z$.
The Memory Model uses a MDRNN to predict the latent vector $z$ that the Vision Model produces by taking the conditional probability $p(z_{t-+1}|a_t, z_t, h_t)$. $a_t$ denotes the action while $h_t$ denotes the hidden state of the RNN at time $t$.

\begin{figure}[htb]
    \centering
    \includegraphics[width=\columnwidth]{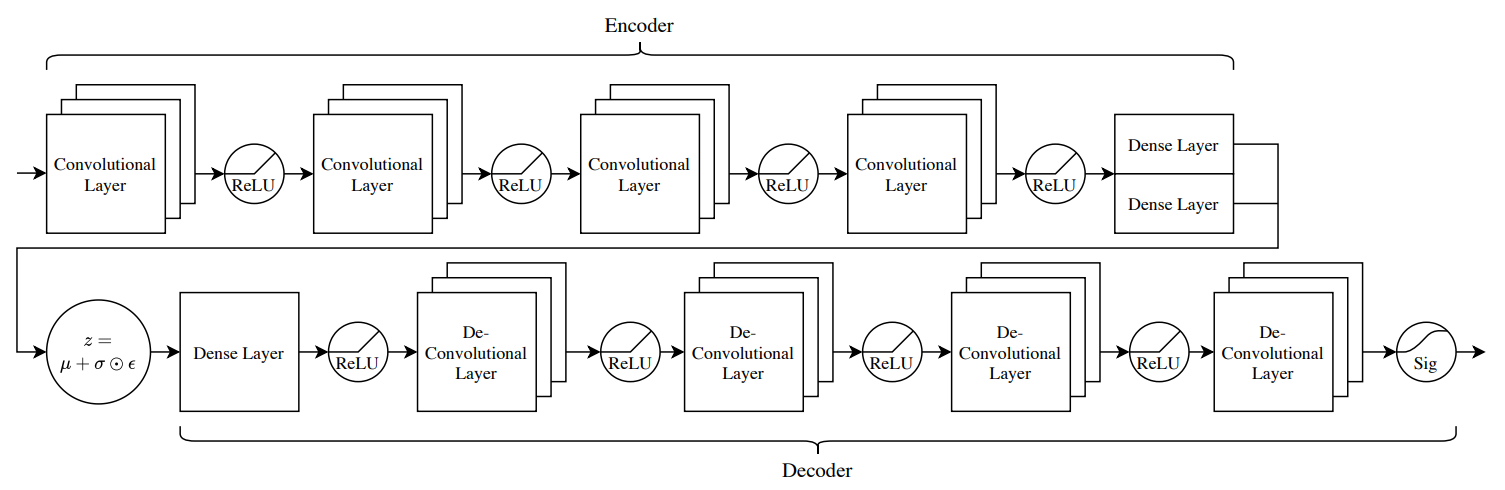}
    \caption{Architecture of the CNN-VAE}
    \label{fig:vae_arch}
\end{figure}
For the Vision Model, a Convolutional VAE (Fig. \ref{fig:vae_arch}) is used.
The input takes a $64\times64$ RGB-frame and transforms it in to a 64 dimensional latent vector $z$.

The encoder part consists of four convolutional layers, each with a ReLU as activation function.
The parameters $\mu$ and $\sigma$ are each calculated from a dense layer.
Using reparameterization, we get the latent representation $z = \mu + \sigma \odot \epsilon$.
The decoder starts with one dense layer, followed by four deconvolutional layers, also with ReLU as activation functions, ending with a Sigmoid layer.
We used a Huber-Loss \cite{huber1992robust} with $\beta=1$ to calculate reconstruction error of the cost function, as it is less sensitive to outliers than MSE-Loss. 
For regularization, Kullback-Leibler-Divergence is used.
We train this VAE over $100$ epochs with the Adam optimizer, a learning rate of $1e^-3$ and batch size $100$.

\begin{figure}[htb]
    \centering
    \includegraphics[width=\columnwidth]{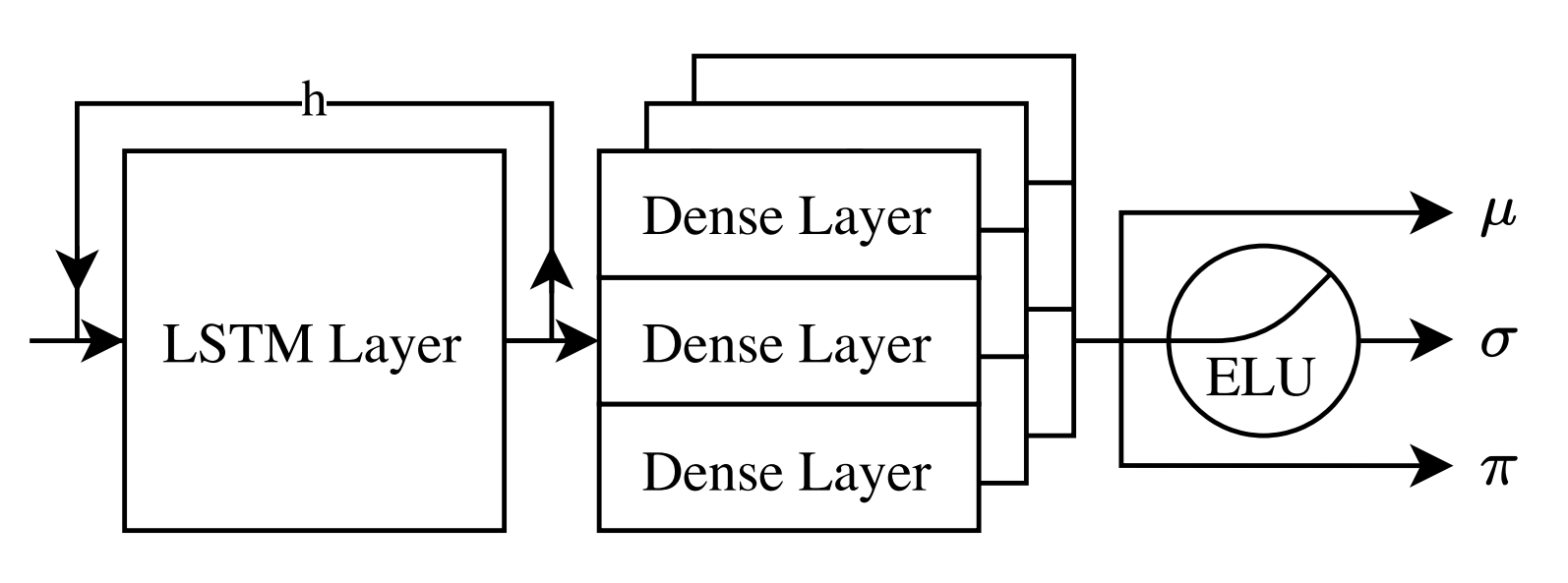}
    \caption{Architecture of the MDRNN}
    \label{fig:rnn_arch}
\end{figure}
The Memory Model uses a MDRNN, consisting of one LSTM layer and three stacked dense layers to model the parameters of the mixing distribution (\autoref{fig:rnn_arch}). It takes an input-sequence of $20$ latent vectors with a dimension of $64$. We trained the MDRNN for $100$ epochs using the negative log likelihood and Huber-Loss with the Adam optimizer, a learning rate of $1e^-3$ and batch size of $20$.

\subsubsection{Evaluation Pipeline}
\label{subsubsec:eval_pipeline}
Inspired by \cite{ha18}, we use a random policy to gather observations as input of a VAE. The dataset consists of $1e5$ $64 \times 64$ RGB-frames generated from selecting random actions [left, right, jump, do nothing]. However, we chose the probabilities of the actions to be not equally distributed [$.15$, $.32$, $.30$, $.50$]. This results in movement towards the goal and avoids jumping repeatedly, which takes 20 timesteps each.
Using the input data as described above, we train a VAE to encode the frames into a 64 dimensional latent space.
Using the trained VAE, another dataset is created in which a single data point is a sequence of 20 encoded observations with corresponding actions and the immediate following observation. However, the encoded observations are not stored as latent representation $z$, but instead as parameters $\mu$ and $\sigma$ from the encoder. They are dynamically reparameterized during the MDRNN training, to prevent overfitting to concrete $z$ values.

The resulting latent dataset is used to train an MDRNN to predict the subsequent state for each sequence of latent observations and actions.
These state transitions are unimodal, but by omitting or masking actions they can be made multimodal. The reason for this is that for a given state, the subsequent state is deterministically defined if the action which the agent performs is known. For example, if the agent chooses the action "right", the environment will be shifted a fixed number of of pixels to the left. However, if the action is unknown, the subsequent state cannot be predicted unambiguously, producing multimodality. Since the dimensionality of the action space is four, there are also four possible different subsequent states. To maintain the size of the input for the MDRNN, the actions are not simply omitted, but replaced by an invalid action. We call this process \textit{masking}.
To obtain both unimodal and multimodal state transitions, actions are masked for the first $50\%$ of the time steps.
For the second half, the randomly chosen valid actions are used.
This way, a benchmarking dataset with known ground-truth (unimodal/multimodal) is created.
Based on this dataset, the $4$ multimodality metrics presented in \autoref{sec:quantify_multimod_mdn} are evaluated.
We further evaluate how the metrics behave for a varying amount $k$ of mixture components.
To do this, $13$ separate MDRNNs with an amount of mixture components varying between $2$ and $50$ are constructed.
In order to factor in the stochasticity of MDRNN training, the complete training process is repeated $10$ times, resulting in a total of $130$ individual models.

\section{\uppercase{Evaluation Results}}

The following section first presents the evaluation results using the inverse sine wave dataset,
followed by the world model based experiment. 

\subsection{Inverse Sine Wave}
\begin{figure}[t]
    \centering
    \begin{subfigure}{.48\columnwidth}
        \includegraphics[width=\textwidth]{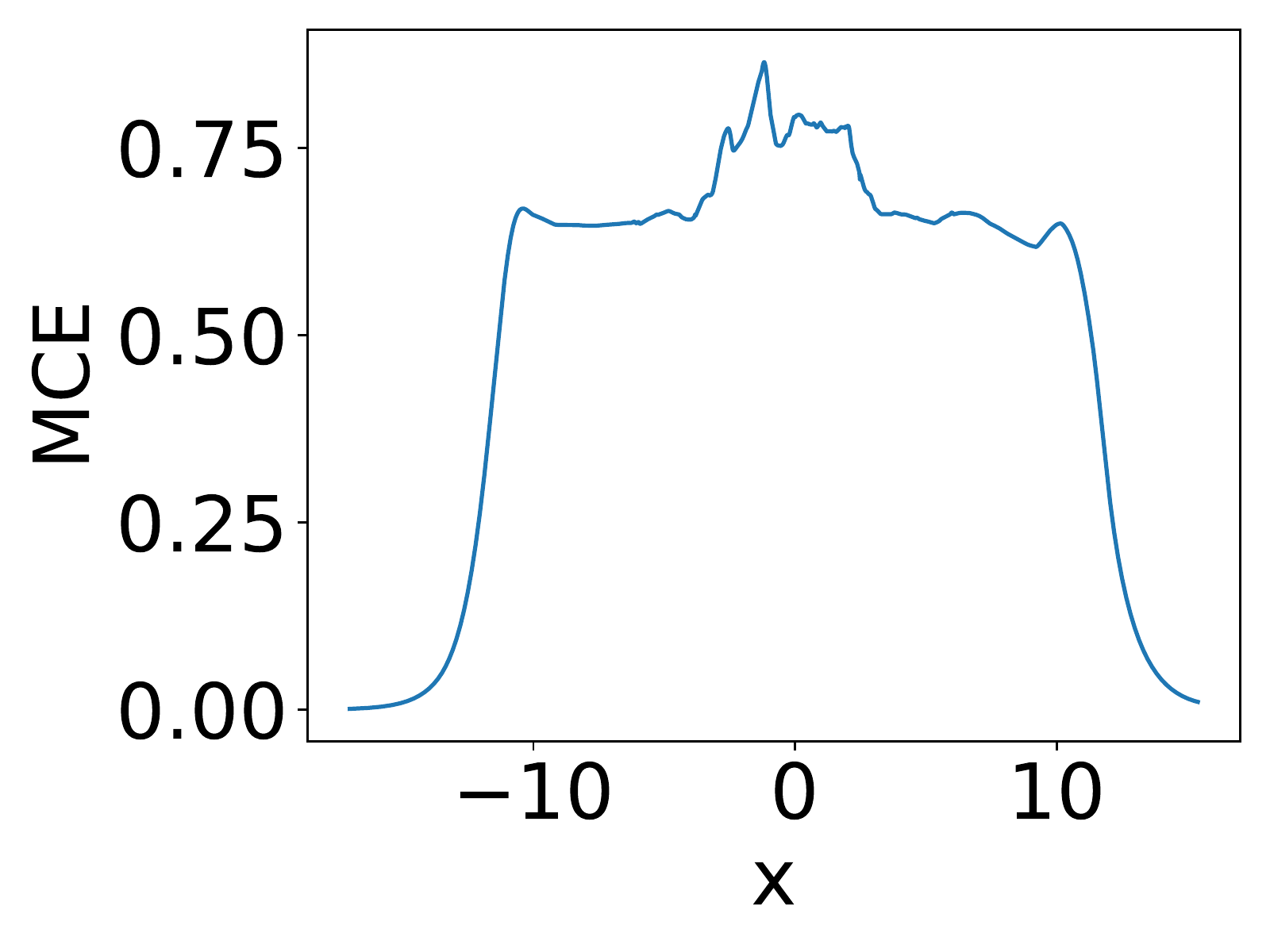}
        \caption{MCE}
    \end{subfigure}
    \begin{subfigure}{.48\columnwidth}
        \includegraphics[width=\textwidth]{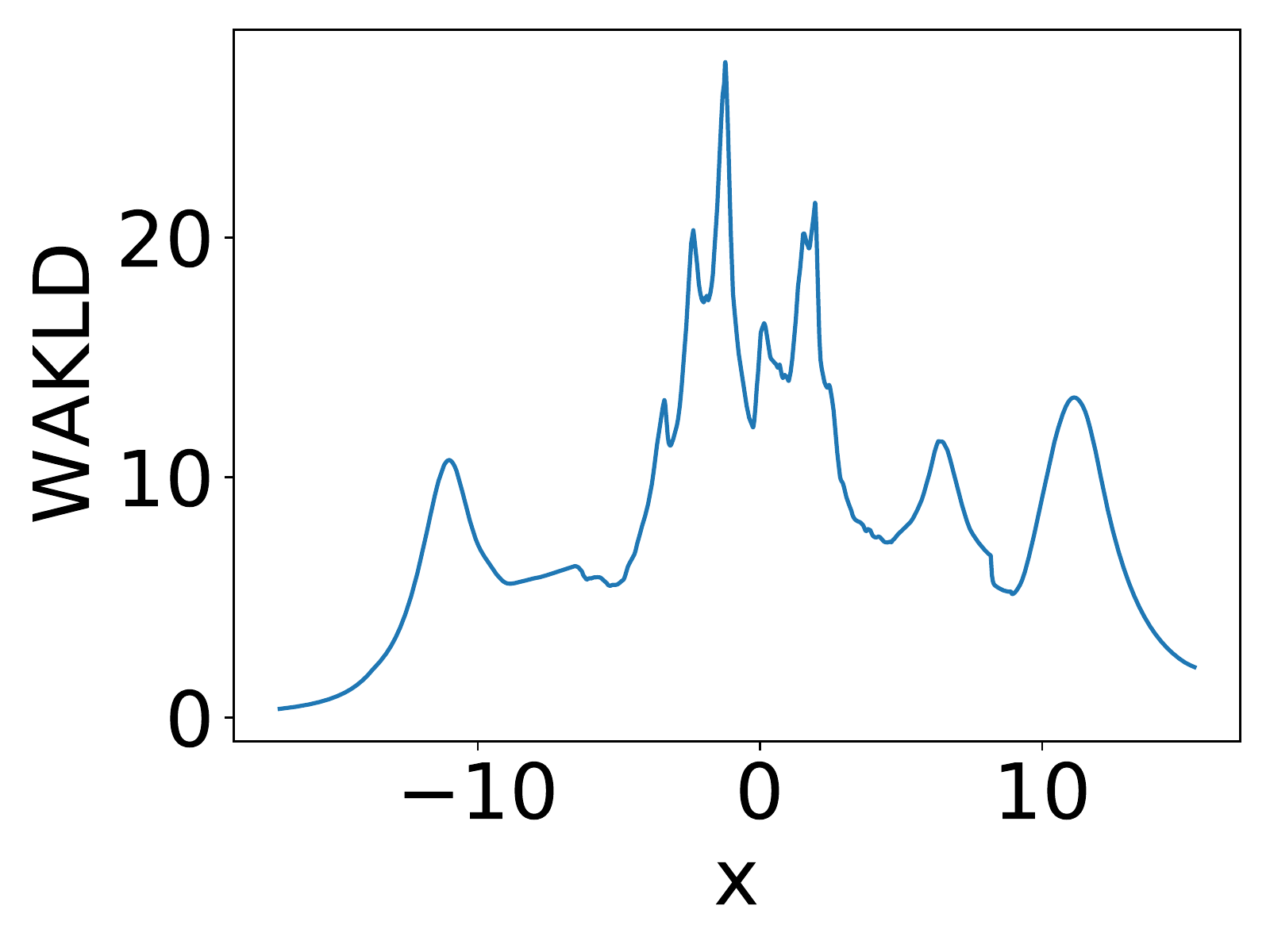}
        \caption{WAKLD}
    \end{subfigure}
    \begin{subfigure}{.48\columnwidth}
        \includegraphics[width=\textwidth]{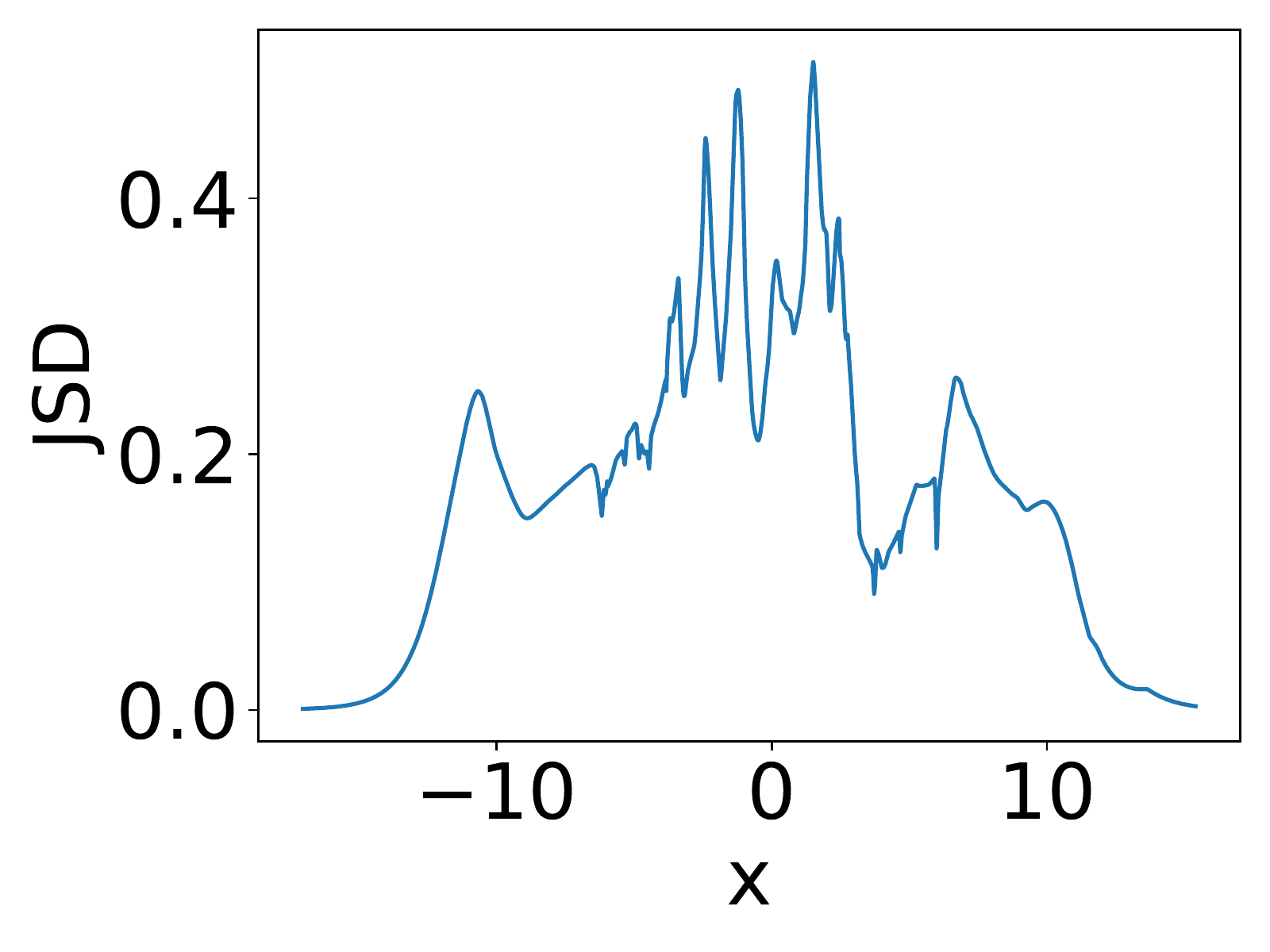}
        \caption{JSD}
    \end{subfigure}
    \begin{subfigure}{.48\columnwidth}
        \includegraphics[width=\textwidth]{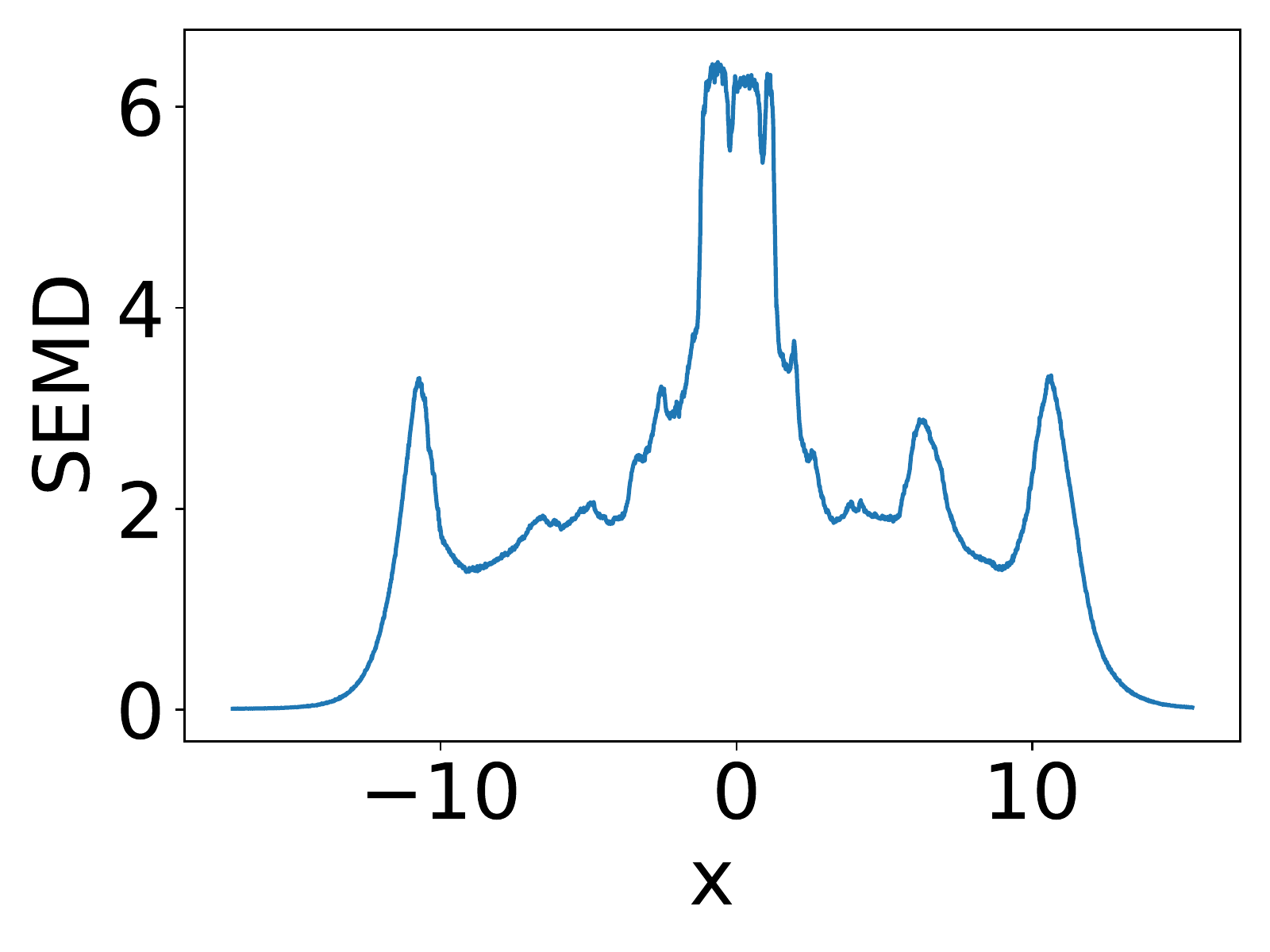}
        \caption{SEMD}
    \end{subfigure}
    \caption{MCE, WAKLD, JSD and SEMD metrics applied to $5$ component MDNs trained on the inverse sine wave dataset. All values shown are averages of $50$ runs.}
    \label{fig:inv_sinus_metrics}
\end{figure}

\autoref{fig:inv_sinus_metrics} shows the evaluation results of applying the $4$ multimodality metrics introduced in \autoref{sec:quantify_multimod_mdn} to MDNs trained on the inverse sine wave dataset.
In order to reduce stochastic effects present when training MDNs, average values of $50$ runs are shown.

What is apparent at a first glance, is that all $4$ metrics behave similarly in the unimodal data range ($-10 > x > 10$ ).
As described in \autoref{sec:quantify_multimod_mdn}, analyzing the mixing coefficients in these areas showed that a single component is predominantely responsible for predicting the values. This unimodality is correctly reflected in all metrics by values converging towards $0$.
Rising values in the intermediate area ($-10 < x < 10$ ) correctly reflect increasing multimodality, peaking around $x=0$, where $3$ maximally separated modes of the inverse sine wave are present.

After having established the basic feasability of the approach, we now evaluate whether the metrics' behaviour transfers to the more complex case of predicting multimodal state-transitions in world models.

\subsection{World Model}

\begin{figure}[t]
    \centering
    \includegraphics[width=\columnwidth]{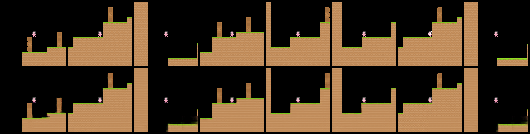}
    \caption{Example of $8$, $64x64x3$ coinrun input states (first row) and their respective VAE reconstructions (second row).}
    \label{fig:vae_coinrun}
\end{figure}

The first step of the evaluation pipeline was the training of the world model's VAE.
Using the $1e5$ samples generated via a random policy, the VAE converged after $100$ epochs and was able to successfully compress and reconstruct the state input. \autoref{fig:vae_coinrun} shows examples of the $64x64x3$ sized original coinrun states in the first row, followed by their respective VAE reconstruction in the second row.
Using this VAE, $13$ separate MDRNNs with between $2$ and $50$ mixture components were constructed.
Training was then repeated $10$ times, resulting in a total of $130$ individual models.

\begin{figure}[t]
    \centering
    \begin{subfigure}{.48\columnwidth}
        \includegraphics[width=\textwidth]{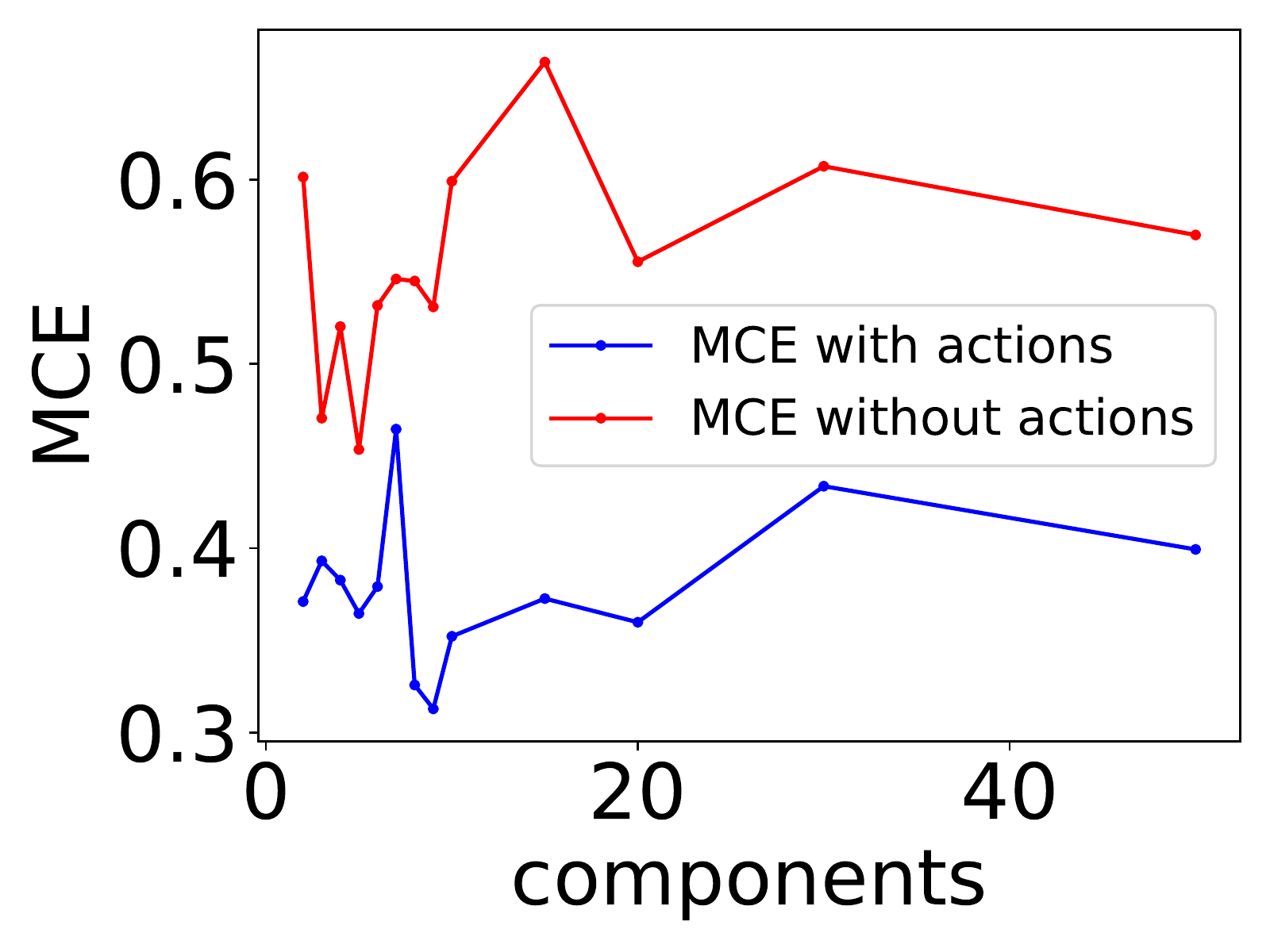}
        \caption{MCE}
    \end{subfigure}
    \begin{subfigure}{.48\columnwidth}
        \includegraphics[width=\textwidth]{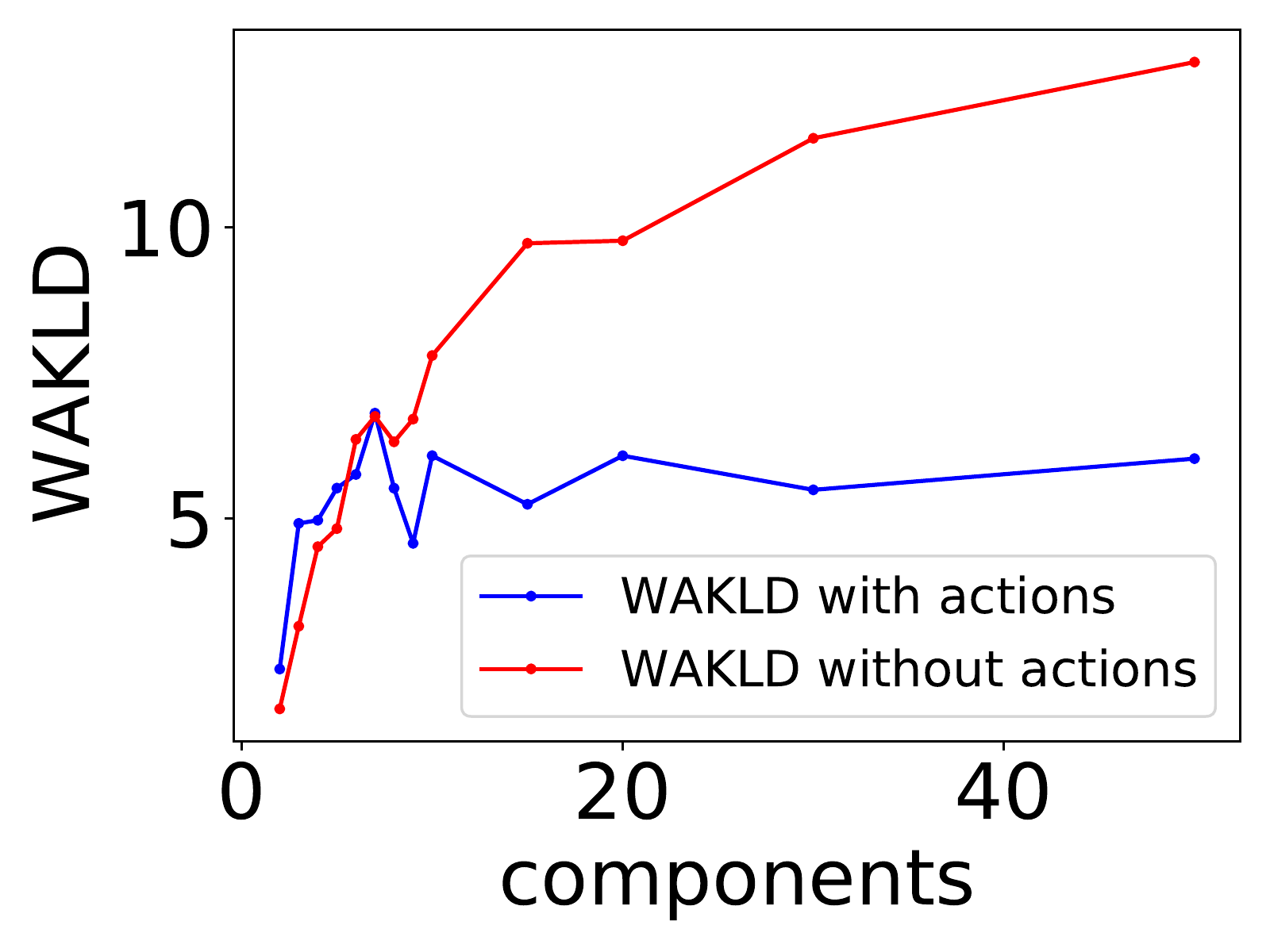}
        \caption{WAKLD}
    \end{subfigure}
    \begin{subfigure}{.48\columnwidth}
        \includegraphics[width=\textwidth]{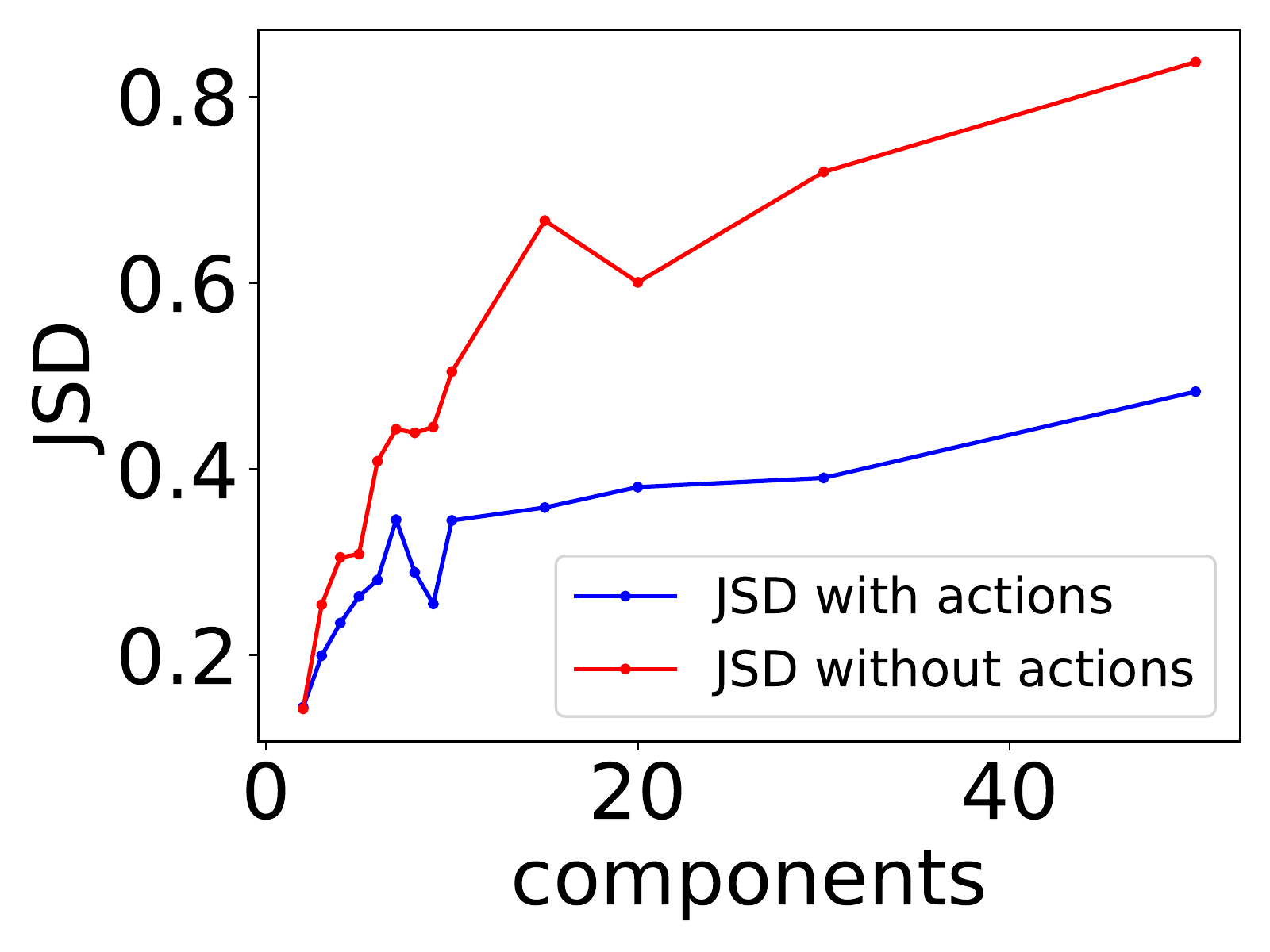}
        \caption{JSD}
    \end{subfigure}
    \begin{subfigure}{.48\columnwidth}
        \includegraphics[width=\textwidth]{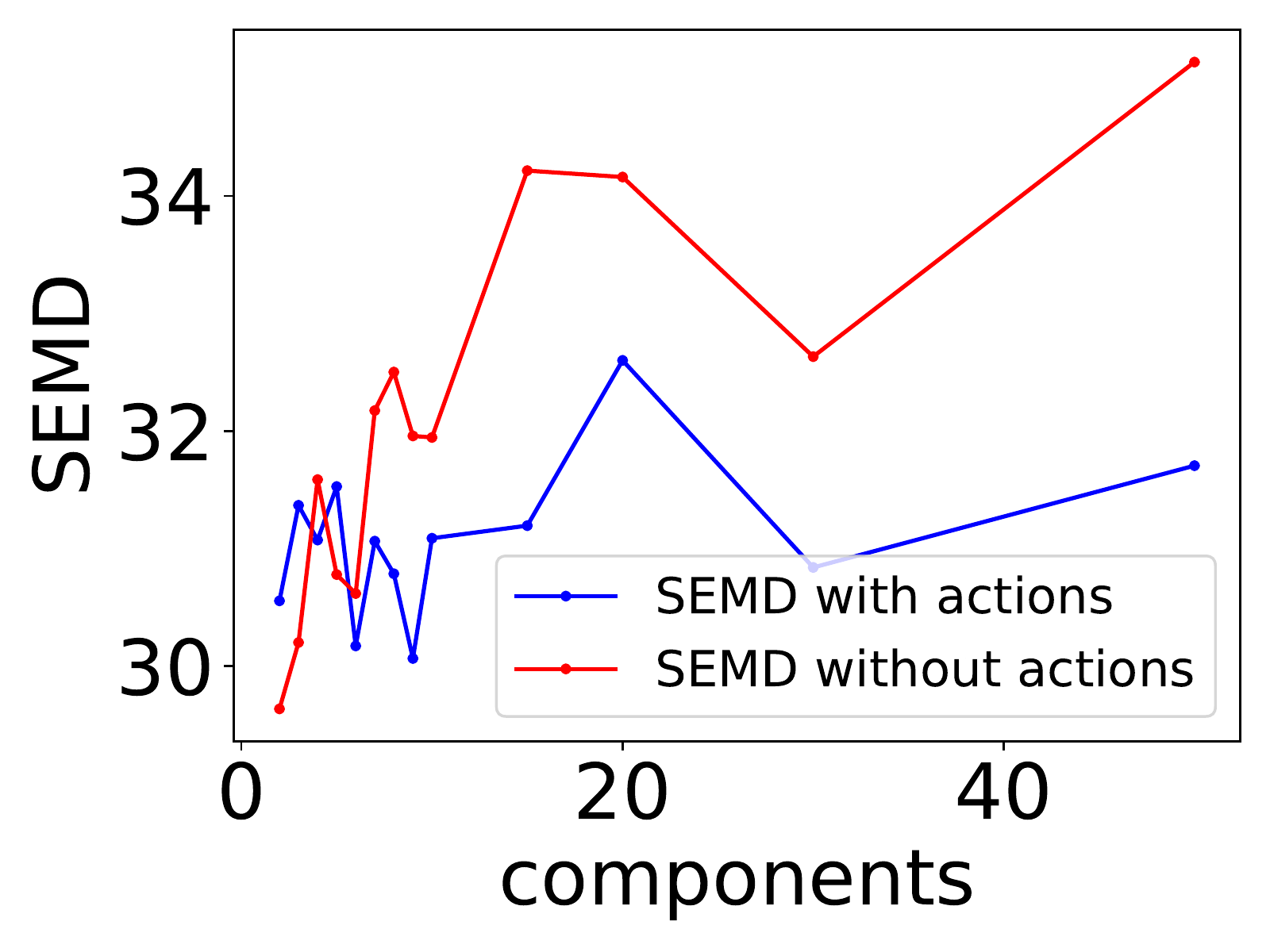}
        \caption{SEMD}
    \end{subfigure}
    \caption{Comparison of the multimodal uncertainty metrics. 
    Blue curves show computed values on the unimodal data, while red curves show values of the multimodal data. All values are averages of $10$ evaluation runs.}
    \label{fig:eval_metrics}
\end{figure}

\autoref{fig:eval_metrics} shows a comparison of the $4$ multimodality metrics when applied to state-transitions with masked actions, i.e. multimodality (red curve) and non-masked actions, i.e. unimodality (blue curve).
All values shown are averages of $10$ evaluation runs.
At a first glance, it is already apparent that all $4$ metrics MCE, WAKLD, JSD and SEMD show the expected behaviour:
Lower metric values in the case of unimodality and higher values in the case of multimodality.
Note here, that it is not possible to compare the absolute values between the different metrics.
What is of interest, from the perspective of building a reliable multimodality detector, is the distance between unimodal values and multimodal values of a single metric. If there is no overlap, and reported values of multimodal data are consistently above values of unimodal data, a reliable differentiation is possible.
For the WAKLD and JSD metrics, the distance of unimodal and multimodal values increases along the number $k$ of mixture components (X-Axis), with only minor drops.
The MCE and SEMD metrics are not as consistent when using a low component count. Here, the reported multimodality values (red curves in \autoref{fig:eval_metrics}) fluctuate strongly for component counts of $k<10$.
In the case of the MCE metric, further increasing the number of components does not increase the multimodality value as much, when compoared to the other metrics.
This results in a more or less constant distance between the unimodal and multimodal curve for $k>10$.

Concerning the overlap of unimodal and multimodal metric values, the MCE and JSD metrics perform best.
Here, no overlap of the curves is present for any number of components used.
The WAKLD and SEMD metrics do not perform as well here.
For WAKLD, the multimodal values only rise above the unimodal ones for a component count $k>6$.
For SEMD, the metric behaviour is strongly fluctuating, and multimodal values only reliably lie above the unimodal ones for a component count $k>5$.
In the cases where the curves overlap, it would not be possible to reliably differentiate multimodality from unimodality. The consequence of this is that it would not be possible to construct a reliable multimodality detector based on a combination of MDN networks of this size and SEMD or WAKLD for multimodality quantification.
Overall, it is apparent that using a larger component count leads to a more reliable differentiation for all evaluated metrics.

\section{\uppercase{Discussion}}

In this work, we presented a first approach for tackling the challenge of detecting multimodality in world models.
As model based reinforcement learning increasingly gains practical relevance, not least through the development of methods like world models, approaches and metrics like the ones evaluated in this work, become of high relevance for the development of reliable and safe RL systems.
Our evaluation results showed that it is possible to detect multimodal state-transitions and differentiate them from unimodal ones, by applying multimodality metrics on the MDN network of a world model architecture.
The metrics we newly introduced in this work, MCE and WAKLD both performed well, allowing for a reliable differentiation when using a mixture component count $k>6$.
Using the symmetric divergence measure JSD turned out to produce the most consistent differentiation between unimodal and multimodal data for any number of components used.
On the other hand, the application of SEMD, which is based on the Wasserstein metric, needs extra care,
as in cases where a low amount of mixture components is used, the reported values fluctuated strongly.
As a consequence, no reliable multimodality detection would be possible here.

As a next step, we plan to use the developed approach and metrics to construct a complete multimodal state-transition one-class classificator.
It would also be interesting to further develop variants of the Wasserstein based SEMD as well as the WAKLD metric, with the goal of improving the metrics for MDNs with low component count.

\bibliographystyle{apalike}
{\small
\bibliography{main}}

\end{document}